\documentclass[sigconf]{acmart}
\AtBeginDocument{%
  \providecommand\BibTeX{{%
    \normalfont B\kern-0.5em{\scshape i\kern-0.25em b}\kern-0.8em\TeX}}}

\copyrightyear{2021}
\acmYear{2021}
\setcopyright{acmcopyright}\acmConference[SIGIR '21]{Proceedings of the 44th International ACM SIGIR Conference on Research and Development in Information Retrieval}{July 11--15, 2021}{Virtual Event, Canada}
\acmBooktitle{Proceedings of the 44th International ACM SIGIR Conference on Research and Development in Information Retrieval (SIGIR '21), July 11--15, 2021, Virtual Event, Canada}
\acmPrice{15.00}
\acmDOI{10.1145/3404835.3463062}
\acmISBN{978-1-4503-8037-9/21/07}
\usepackage{subcaption}
\usepackage{amsmath}
\usepackage{amsfonts}
\usepackage{multirow}
\usepackage{multicol}
\usepackage{booktabs}
\usepackage{balance}
\usepackage{graphicx}

\DeclareMathOperator{\MASK}{MASK}
\DeclareMathOperator{\relu}{LeakyReLU}
\DeclareMathOperator{\meanp}{mean}
\DeclareMathOperator{\maxp}{max}

\begin{document}


\fancyhead{}

\title{Predicting Patient Readmission Risk from Medical Text via Knowledge Graph Enhanced Multiview Graph Convolution}

%\author{
%Qiuhao Lu\textsuperscript{1}, Thien Huu Nguyen\affmark[1], Dejing Dou\affmark[1]\affmark[2]\\}
%\affiliation{\institution{\affmark[1]Department of Computer and Information Science, University of Oregon}}
%\affiliation{\institution{\affaddr{\affmark[2]Baidu Research}}}
%\email{{luqh, thien, dou}@cs.uoregon.edu, doudejing@baidu.com}
\author{Qiuhao Lu}
\affiliation{\institution{University of Oregon}\city{Eugene}
  \state{OR}\country{USA}}
\email{luqh@cs.uoregon.edu}

\author{Thien Huu Nguyen}
\affiliation{\institution{University of Oregon}\city{Eugene}
  \state{OR}\country{USA}}
\email{thien@cs.uoregon.edu}

\author{Dejing Dou}
\affiliation{\institution{University of Oregon}
\institution{Baidu Research}\city{Eugene}
  \state{OR}\country{USA}}
\email{dou@cs.uoregon.edu}
\email{doudejing@baidu.com}

\begin{abstract}
Unplanned intensive care unit (ICU) readmission rate is an important metric for evaluating the quality of hospital care. Efficient and accurate prediction of ICU readmission risk can not only help prevent patients from inappropriate discharge and potential dangers, but also reduce associated costs of healthcare. In this paper, we propose a new method that uses medical text of Electronic Health Records (EHRs) for prediction, which provides an alternative perspective to previous studies that heavily depend on numerical and time-series features of patients. More specifically, we extract discharge summaries of patients from their EHRs, and represent them with multiview graphs enhanced by an external knowledge graph. Graph convolutional networks are then used for representation learning. Experimental results prove the effectiveness of our method, yielding state-of-the-art performance for this task.
\end{abstract}





\maketitle

\section{Introduction}

Patients who are readmitted to intensive care units (ICUs) after transfer or discharge usually have a greater chance of developing dangerous symptoms that can result in life-threatening situations. Readmissions also put families at higher financial burden and increase healthcare providers' costs. Therefore, it is beneficial for both patients and hospitals to identify patients that are inappropriately or prematurely discharged from ICU.

Over the past few years, there has been a surge of interest in applying machine learning techniques to clinical forecasting tasks, such as readmission prediction \cite{lin2019analysis}, mortality prediction \cite{Harutyunyan2017MultitaskLA}, length of stay prediction \cite{ma2020length}, etc. Earlier studies generally select statistically significant features from patients' Electronic Health Records (EHRs), and feed them into traditional machine learning models like logistic regression \cite{xue2018predicting}. Deep learning models have also been gaining more and more attention in recent years, and have shown superior performance in medical prediction tasks. For example, Lin \textit{et al.} select $17$ types of chart events (diastolic blood pressure, capillary refill rate, etc.) over a 48-hour time window and put them into a LSTM-CNN model \cite{lin2019analysis} and achieve much better performance than previous work in readmission prediction.

\begin{figure*}[t]
    \centering
    \includegraphics[scale=0.27]{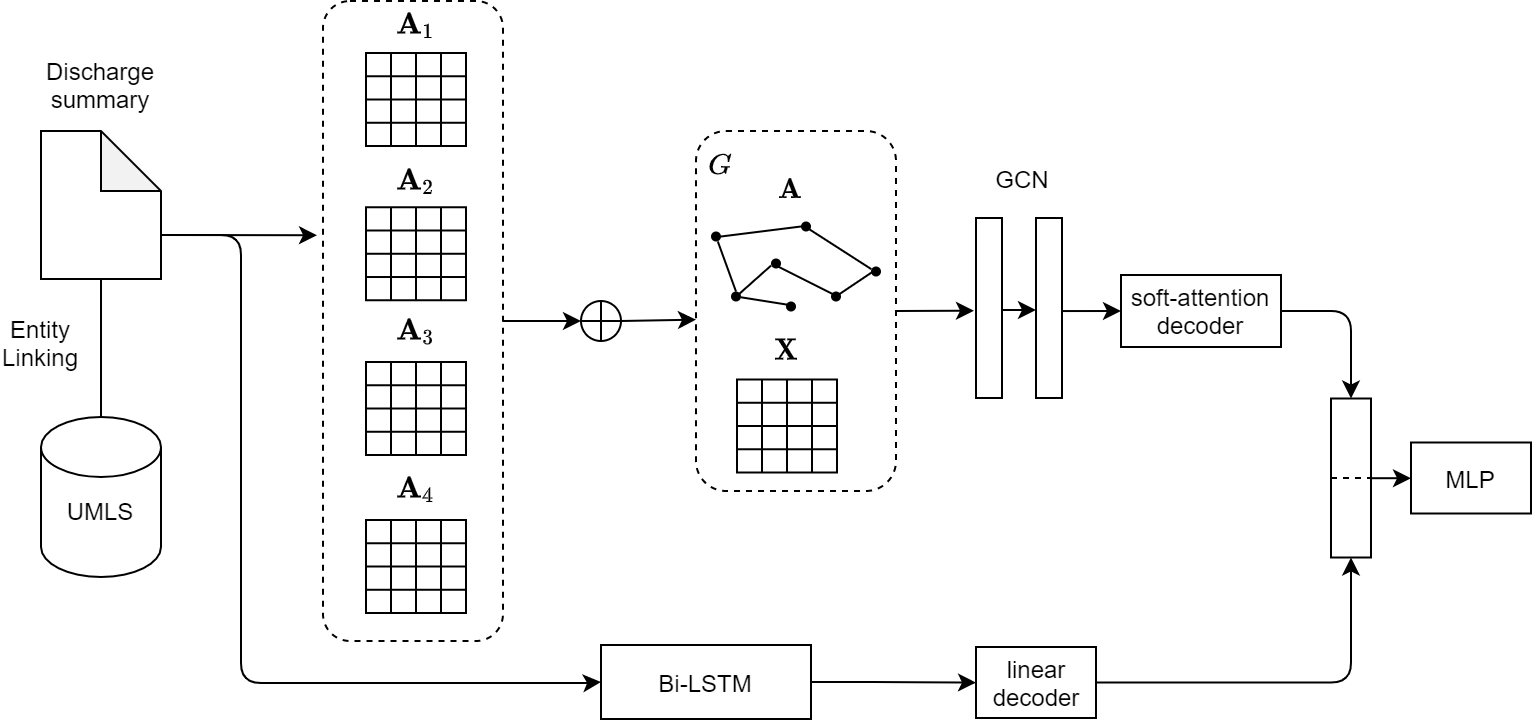}
    \caption{Architecture of MedText.}
    \label{framework}
\end{figure*}

A common theme among these studies is that they all rely on numerical and time-series features of patients, while neglecting rich information in the clinical notes of EHRs. This motivates us to tackle this task from a pure natural language processing perspective, which is not well explored in literature. Essentially, in this work, we consider the task of ICU readmission prediction as binary text classification, i.e., for a given clinical note, the model aims to predict whether or not the patient will be readmitted to ICU within 30 days after discharge.

Although it is possible to directly apply existing text classification methods to the readmission prediction task, two major challenges need to be addressed: (1) clinical notes, e.g., discharge summaries, are generally long and noisy, which makes it difficult to capture the inherent semantics to support classification; (2) general methods do not consider domain knowledge in the medical area, which is critical as medical concepts are hard to interpret with limited training for downstream tasks.

Recently, a useful strategy is proposed to tackle the first challenge, where it encodes documents with graphs-of-words to enhance the interactions of context, and to capture the global semantics of the document. The strategy has been applied to different NLP tasks, including document-level relation extraction \cite{christopoulou2019connecting,nan2020reasoning,chen2020dialogue}, question answering \cite{de2019question,qiu2019dynamically}, and text classification \cite{yao2019graph,zhang2020every,nikolentzos2020message}. But constructing graphs of clinical notes for patient outcome prediction, to our knowledge, is underexplored.

Motivated by this, we propose a novel graph-based model that represents clinical notes as document-level graphs to predict patient readmission risk. Moreover, to address the second challenge, we incorporate an external knowledge graph, i.e., the Unified Medical Language System (UMLS) \cite{bodenreider2004unified} Metathesaurus, to construct a four-view graph for each input clinical note. The four views correspond to intra-document, intra-UMLS, and document-UMLS interactions, respectively. By constructing such a enhanced graph representation for clinical notes, we inject medical domain knowledge to improve representation learning for the model. Our contribution can thus be summarized as follows:

%We summarize our contributions as follows:

\begin{itemize}
  \item We propose a novel graph-based text classification model, i.e., MedText, to predict ICU patient readmission risk from clinical notes in patients' EHRs. Unlike previous studies that rely on numerical and time-series features, we only use clinical notes to make predictions, which provides some insights on utilizing medical text for clinical predictive tasks.
  %\item We propose a novel graph-based text classification model, i.e., MedText, to predict ICU patient readmission risk, using clinical notes from patients' EHRs. We construct a specifically designed multiview graph for each clinical note to capture the intra-document, intra-UMLS, and document-UMLS interactions among words and entities.
  %\item We incorporate an external knowledge graph, i.e., UMLS, to generate an specifically designed four-view graph representation of clinical notes, and encode them with the proposed model. The experimental studies demonstrate the superb performance of this method, by updating the state-of-the-art results on readmission prediction.
  \item We construct a specifically designed multiview graph for each clinical note to capture the interactions among words and medical concepts. In this way we inject domain-specific information from an external knowledge graph, i.e., UMLS, into the model. The experimental studies demonstrate the superb performance of this method, by updating the state-of-the-art results on readmission prediction.
\end{itemize}

\section{Methodology}

\subsection{Graph Construction}
For each document (e.g., clinical note), we construct a weighted and undirected four-view graph $\mathcal{G} = (\mathcal{N},\mathcal{E})$ with an associated adjacency matrix $\mathbf{A}$, where $\mathcal{N}$ and $\mathcal{E}$ refer to the vertex set and edge set respectively. We also denote the representation of vertices by $\mathbf{X}$. Instead of using unique words in the document as vertices, we first conduct entity linking over the text and link the entity mentions to UMLS\footnote{We use ScispaCy \cite{neumann-etal-2019-scispacy} as the entity linker in this work.}. Consequently, we consider two types of vertices in the document-level graph $\mathcal{G}$, i.e., the unique words $\mathcal{N}_w$ and the linked UMLS entities $\mathcal{N}_e$. The vertex set $\mathcal{N}$ is thus formed as the union of $\mathcal{N}_w$ and $\mathcal{N}_e$: $\mathcal{N}=\mathcal{N}_w \cup \mathcal{N}_e$. Four views are then designed to exploit intra-document, intra-UMLS, and document-UMLS interactions that will be combined to form the adjacency matrix as follows.

\subsubsection{Intra-Document: $\mathcal{V}_1$}
$\mathcal{V}_1$ is designed to capture the intra-document interactions among words and entities. Essentially, we expect the edge weights between vertices to estimate the level of interaction, so that vertices can directly interact during message passing even if they are sequentially far away from each other in the document. In this work, we generate the adjacency matrix $\mathbf{A}_1$ for $\mathcal{V}_1$ by counting the co-occurrences of vertices within a fixed-size sliding window (size 3 in this work) over the text.

\subsubsection{Intra-UMLS: $\mathcal{V}_2$,$\mathcal{V}_3$}
In this work, we aim to inject external knowledge from UMLS to the document-level graph representation. To this end, we consider two types of information, i.e., the internal structure of UMLS and the semantic similarities between medical concepts. Specifically, we construct $\mathcal{V}_2$ by computing the shortest path lengths between entity vertices as edge weights in $\mathbf{A}_2$, where a shorter path indicates a higher relevance. We further construct $\mathcal{V}_3$ by computing the string similarities based on the word overlap ratios of entity descriptions for $\mathbf{A}_3$.

\subsubsection{Document-UMLS: $\mathcal{V}_4$}
$\mathcal{V}_4$ is constructed by calculating the cosine similarities between initial representations of all vertices, including words and entities, which aims to capture the interactions between the information sources, i.e., the document itself and the knowledge base. The similaries are used for edge weights $\mathbf{A}_4$.

\subsubsection{View Combination}
By combining the four views, we expect to leverage three levels of interactions, i.e., intra-document, intra-UMLS, and document-UMLS, to generate rich interaction structures for documents to aid representation learning. Intuitively, the four views are combined via a weighted sum of the four adjacency matrices as the final adjacency matrix $\mathbf{A}$:

\begin{equation}
    \mathbf{A} = \MASK (\sum_{i=1}^{4} \alpha_i \mathbf{A}_i)
    \label{finalview}
\end{equation}
where $\mathbf{A}_i$ refer to each view's normalized adjacency matrix and $\alpha_i$ are the balancing factors that are determined by cross-validation. The adjacency matrix is then masked with a threshold, i.e., $\gamma=0.5$, where only edges with larger weights are kept for further message passing. The motivation for the masking is to improve robustness and efficiency by decreasing some density.

%\begin{table*}[t]
%\begin{center}
%\begin{tabular}{lrrrrrrr}
%    %\hline
%    \toprule
%    Method	& Acc & Pre-0 & Pre-1 & Re-0 & Re-1 & \textbf{A.R} & A.P \\
%    \midrule
%    \cite{lin2019analysis}  & 0.698 &0.916 &0.367 &0.687 &0.742 &0.791 &0.513 \\
%    \midrule
%    LSTM  &0.840&	0.859&	0.704&	0.956&	0.366&	0.794&	0.600\\
%    \midrule
%    CC-LSTM  & 0.848&	0.854&	0.786&	0.978& 0.321& 0.804& 0.613\\
%    \midrule
%    MedText & 0.850&0.858 & 0.768 & 0.975 & 0.340 & \textbf{0.824} & 0.631\\
%    \bottomrule
%\label{table1}
%\end{tabular}

%\footnotesize
%*Acc: Accuracy, Pre: Precision, Re: Recall, A.R: AUC under ROC, A.P: AUC under PRC
%\end{center}
%\caption{Performance on 30-day unplanned ICU patient readmission prediction.}
%\label{table1}
%\end{table*}

The representation of vertices, i.e., $\mathbf{X}$, are initialized with a pre-trained word embedding BioWordVec \cite{zhang2019biowordvec}. For entity vertices, we take the average values of word embeddings of the entity names as the representation for the entity.

\subsection{Encoding and Decoding}

In this work, we incorporate a two-layer graph convolutional network (GCN) \cite{kipf2016semi} to encode the graph representation of clinical notes, as depicted in Figure~\ref{framework}. We include an attention layer after GCN, which serves as a decoder to decode the document-level representation $\mathbf{D}_G$ from node embeddings. The encoding process can be described as:
\begin{equation}
    \mathbf{X}^{(l+1)} = \relu (\mathbf{\hat{D}}^{-\frac{1}{2}} \mathbf{\hat{A}}\mathbf{\hat{D}}^{\frac{1}{2}} \mathbf{X}^{(l)} \mathbf{W}^{(l)})
\end{equation}
where $\mathbf{\hat{A}}=\mathbf{A}+\mathbf{I}$, and $\mathbf{I}$ is the identity matrix of $\mathbf{A}$. $\mathbf{\hat{D}}$ is the diagonal degree matrix of $\mathbf{\hat{A}}$, and $\mathbf{W}^{(l)}$ is the weight matrix for the $l$-th layer where $l=0,1,2$ in this work. 

We incorporate a graph summation module \cite{li2015gated,zhang2020every} to decode the document-level representation $\mathbf{D}_G$ from the constructed graph, by assigning different attention weights to the nodes. The decoding process can be described as:
\begin{equation}
    \mathbf{X}_G = f_1(\mathbf{X}^{(2)}) \odot f_2(\mathbf{X}^{(2)})
\end{equation}
\begin{equation}
    \mathbf{D}_G = \meanp(\mathbf{X}_G) + \maxp(\mathbf{X}_G)
\end{equation}
where $\mathbf{X}^{(2)}$ is the output of the GCN encoder and $f_1$, $f_2$ are two feed-forward networks with sigmoid and leakyrelu activation, respectively. The $f_1$ network acts as a soft attention mechanism that indicates the relative importance of nodes, while $f_2$ serves as feature transformation. The operator $\odot$ denotes element-wise multiplication. Then the document-level representation $\mathbf{D}_G$ is summarized as the addition of the mean and maximum values of the attentive node embeddings.

We also use a two-layer bidirectional LSTM to directly encode the document and decode the document-level representation $\mathbf{D}_T$ with a linear decoder, where linear transformation and max-pooling are applied. Then the two document-level representations, i.e., $\mathbf{D}_G$ and $\mathbf{D}_T$, are concatenated and fed into a MLP classifier. The model is optimized with cross-entropy loss.

\section{Experiment}

\subsection{Dataset}
The experiment is conducted based on the MIMIC-III Critical Care (Medical Information Mart for Intensive Care III) Database, which is a large, freely-available database composed of de-identified EHR data \cite{johnson2016mimic}. For a fair comparison, we use the same data split with the baseline \cite{zhang2020learning}, where the discharge summaries are extracted from EHRs and the generated $48,393$ documents are split into training ($80\%$), validation ($10\%$), and testing ($10\%$).%\footnote{The code will be released upon acceptance.}

%\subsection{Cohort}
%In this experiment, we consider the following $4$ types of ICU readmissions, i.e., the patients who were transferred to low-level wards from ICU and readmitted to ICU within 30 days; the patients who were transferred out of ICU and died later; the patients who were discharged and readmitted to ICU later; and the patients who were discharged and died later. Note that the ``later'' here means ``within 30 days.''

\subsection{Evaluation Metrics}
We use three metrics for evaluation, i.e., the area under the receiver operating characteristics curve (AUROC), the area under the precision recall curve (AUPRC), and the recall at precision of $80\%$ (RP80). AUROC and AUPRC are widely used for evaluating patient outcome prediction tasks, including readmission prediction \cite{zhang2020learning,lu2019learning,lin2019analysis}. RP80 is a clinically-relevant metric that helps minimize the risk of alarm fatigue, as introduced in ClinicalBERT \cite{huang2019clinicalbert}, where we fix the precision at $80\%$ and calculate the recall rate.

\subsection{Baselines}
The following baselines are used for comparison.
\begin{itemize}
    %\item Lin \textit{et al.} \cite{lin2019analysis} presents
    \item BioBERT. BioBERT is a domain-specific BERT variant pre-trained on large biomedical corpora, e.g., PubMed abstracts and PMC full-text articles \cite{lee2020biobert}. In the experiment, we use the latest version, i.e., BioBERT v1.1, with a classification head as the baseline. The last $512$ tokens of each note are used as input to the model.
    \item ClinicalBERT. ClinicalBERT is initialized from BioBERT v1.0 and pre-trained on MIMIC notes \cite{alsentzer-etal-2019-publicly}. Note that there is another ClinicalBERT \cite{huang2019clinicalbert} model which presents a similar idea.
    \item CC-LSTM. Zhang \textit{et al.} propose CC-LSTM that encodes UMLS knowledge into text representations and report state-of-the-art performance on readmission prediction on the MIMIC-III dataset \cite{zhang2020learning}. For a fair comparison, we use the same pre-trained word embeddings, i.e., BioWordVec \cite{zhang2019biowordvec}, in our model. 
    \item MedText-x. Specifically, we replace the Bi-LSTM encoder with ClinicalBERT and BioBERT to demonstrate the effectiveness of the proposed graph-based knowledge injection strategy. The last two baselines are denoted by MedText-ClinicalBERT and MedText-BioBERT, respectively.
\end{itemize}

\subsection{Results}
%The experimental results are presented in Table~\ref{table1}. LSTM-CNN reports the state-of-the-art performance with numerical features as input \cite{lin2019analysis}. CC-LSTM reports the state-of-the-art performance with textual data as input \cite{zhang2020learning}, and LSTM is a simple baseline with an identical classification head with CC-LSTM. The latter two methods are comparable as they only use the discharge summaries for prediction.

%Note that in the readmission prediction task, the Area Under the Receiver Operating Characteristics curve (AUROC, or A.R) is considered as the primary metric for evaluation. In Table~\ref{table1}, along with A.R, some more metrics are proposed for additional evaluation.

The experimental results are presented in Table~\ref{table1}. Generally, the proposed method, i.e., MedText, compares favorably with all the other baselines and outperforms the state-of-the-art method. Besides, directly applying pre-trained language models, such as BioBERT and ClinicalBERT, to readmission prediction does not work well. It is most likely due to the long and noisy nature of clinical notes, and only the last $512$ tokens are taken as input in the experiment. However, by combining with MedText, the performance gets improved greatly, indicating the effectiveness of the proposed graph-based knowledge injection method.

%\begin{table}[t]
%\begin{center}
%\resizebox{.48\textwidth}{!}{\begin{tabular}{lrrrrrr}
%    %\hline
%    \toprule
%    Method & Type&Acc & F1 & \textbf{A.R} & A.P \\
%    \midrule
%    LSTM-CNN & N& 0.698 &0.491 &0.791 &0.513 \\
%    \midrule
%    LSTM &T &0.840&	0.482&	0.794&	0.600\\
%    \midrule
%    CC-LSTM &T  & \textbf{0.848}&	0.456& 0.804& 0.613\\
%    \midrule
%    %MedText & 0.850&0.858 & 0.768 & 0.975 & 0.340 & \textbf{0.824} & 0.631\\
%    MedText &T & 0.847&\textbf{0.493} & \textbf{0.825} & \textbf{0.632}\\
%    \bottomrule
%\label{table1}
%\end{tabular}}

%\footnotesize
%*Acc: Accuracy, N: Numerical, T: Textual, F1: F1 score, A.R: AUC under ROC, A.P: AUC under PRC
%\end{center}
%\caption{Performance on 30-day unplanned ICU patient readmission prediction.}
%\label{table1}
%\end{table}

\begin{table}[t]
\begin{center}
\caption{Performance on 30-day unplanned ICU patient readmission prediction.}
\label{table1}
\begin{tabular}{lccc}
    %\hline
    \toprule
    Method  & AUROC & AUPRC & RP80 \\
    \midrule
    %Lin \textit{et al.} \cite{lin2019analysis}  &0.791 &0.513 & N/A \\
    BioBERT   &	0.775&	0.538 & 0.200\\
    MedText-BioBERT  &	0.811&	0.610 & 0.278\\
    \midrule
    ClinicalBERT   &0.781&	0.536 & 0.189\\
    MedText-ClinicalBERT  &	0.812&	0.615 & 0.277\\
    \midrule
    CC-LSTM \cite{zhang2020learning} & 0.804& 0.613 & N/A\\
    MedText & \textbf{0.825} & \textbf{0.632} & \textbf{0.319}\\
    \bottomrule
\end{tabular}
\end{center}
\end{table}

Additionally, Lin \textit{et al.} propose a readmission prediction model that takes numerical features, e.g., chart events, of patients as input, and claim a state-of-the-art AUROC of $0.791$ with AUPRC of $0.513$ on the same dataset \cite{lin2019analysis}. This is essentially not comparable as they are using numerical features instead of text, but it highlights the value of clinical notes in EHRs.

%which is essentially not comparable with the other three methods. But they report the state-of-the-art performance with numerical features, which actually highlights the value of textual data in EHRs. 

%There is also a precision-recall balance pattern, where textual-based methods tend to have better precision \footnote{We confirmed the numbers with the author.} and numerical-based methods show better recall. 

%Similar patterns can be found in experiments of a recent medical language model, ClinicalBERT \cite{huang2019clinicalbert}. We leave this for further study.

\begin{table}[t]
\begin{center}
\caption{Ablation analysis of MedText.}
\label{table2}
\begin{tabular}{lccc}
    %\hline
    \toprule
    Method	 & AUROC & AUPRC & RP80 \\
    \midrule
    w/o $\mathcal{V}_1$  & 0.803 &0.605 &0.300 \\
    w/o $\mathcal{V}_{1,2}$  &0.809	&0.615&	0.296\\
    w/o $\mathcal{V}_{1,2,3}$ &0.801	&0.607&	0.290\\
    w/o $\mathcal{V}_{1,2,3,4}$ &0.799 &0.601&	0.288\\
    \midrule
    w/o $\mathbf{D}_T$  &	0.808&0.601&	0.275\\
    \midrule
    Full &    0.825 & 0.632 & 0.319\\
    \bottomrule
\end{tabular}
\end{center}
\end{table}

\subsection{Ablation and Sensitivity Study}
\begin{figure}[t]
    \centering
    \includegraphics[scale=0.55]{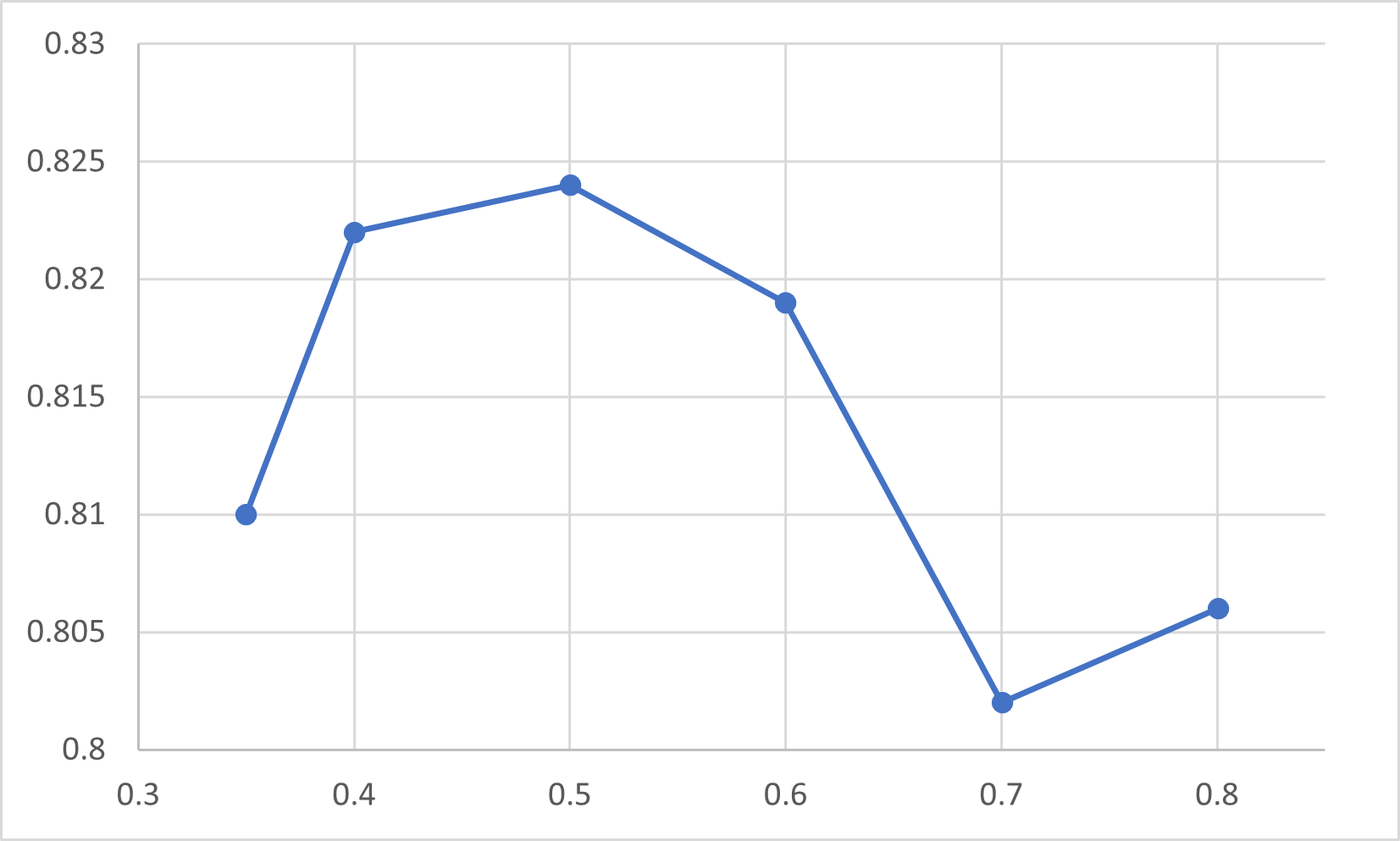}
    \caption{Sensitivity of masking threshold $\gamma$.}
    \label{masking}
\end{figure}

\begin{figure}[t]
    \centering
    \includegraphics[scale=0.6]{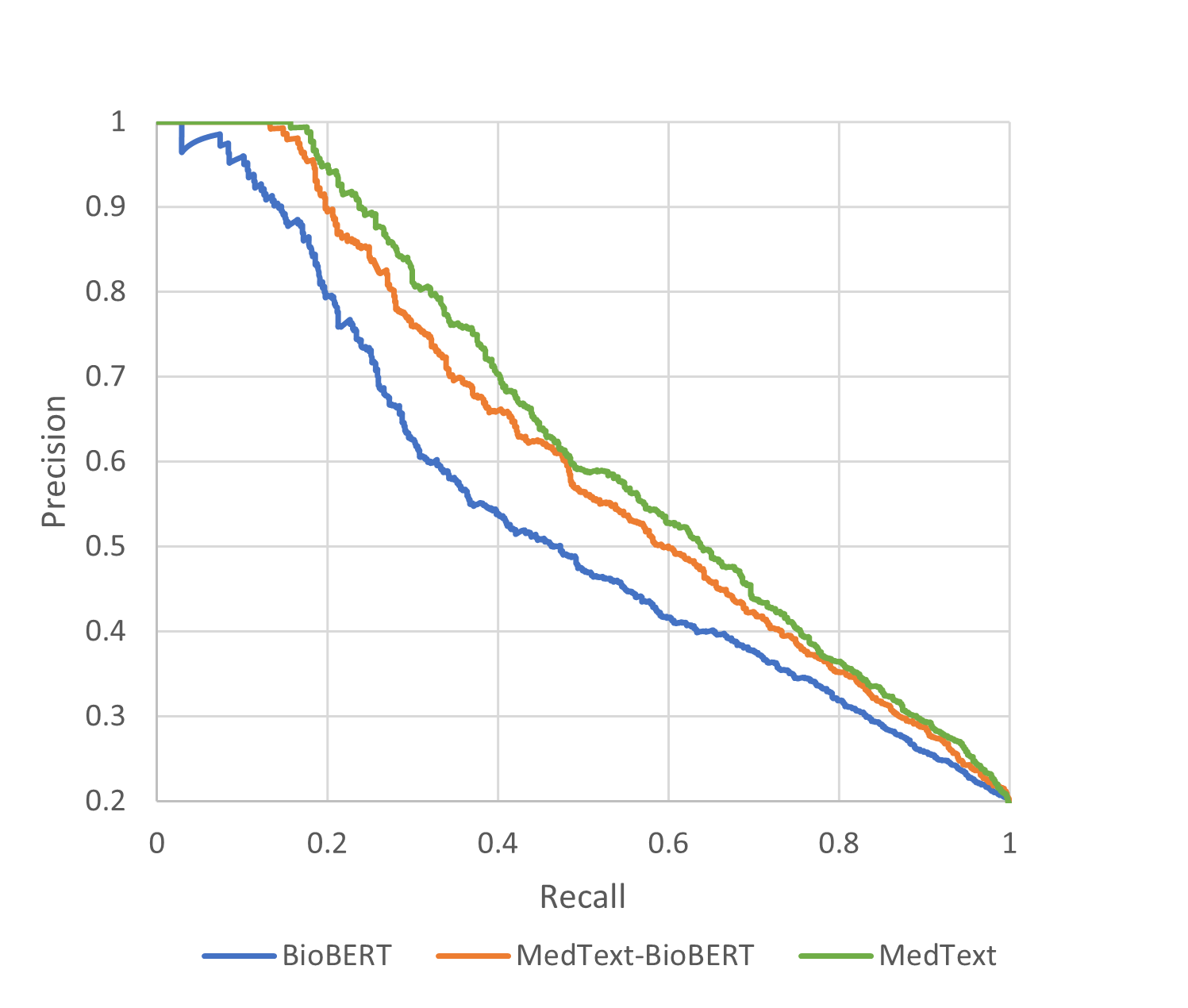}
    \caption{Precision-recall curve of MedText.}
    \label{prcurve}
\end{figure}

We present the ablation study in Table~\ref{table2}. As shown in the table, removal of the four views will cause the performance to drop greatly, indicating the effectiveness and necessity of the four views. It is also worth noticing that the model still performs on par with CC-LSTM if the Bi-LSTM module is removed, i.e., w/o $\mathbf{D}_T$, and it would be more efficient in training. We also show the AUROC score with different masking threshold in Figure~\ref{masking}, where AUROC reaches the peak when $\gamma=0.5$. To further assess the performance of the model in terms of precision and recall, we show the P-R curve in Figure \ref{prcurve}.

\subsection{Error Analysis}
%As in Table~\ref{table2}, we notice that the recall rate is lower than the precision rate (calculated at a default probability threshold $0.5$), which indicates a higher false negative rate, meaning patients with severe conditions (thus likely to be readmitted) are not precisely recognized. It can be caused by entity linking errors, as all four views either directly or indirectly depend on the linked entities. After manually examining a subset of notes, we roughly estimate that $15\%$ to $25\%$ of entities are not recognized or linked, which may have negatively influenced the prediction model. Here are some snippets of a clinical note:

%By setting appropriate parameters for the ScispaCy linker, high precision can be achieved, but at the cost of relatively low recall. 

Entity linking plays an important role in this method as it is the first step of graph construction and all four views either directly or indirectly depend on the linked entities. Since a relatively high linking precision can be achieved by setting appropriate parameters of the ScispaCy linker, we mainly focus on the missed entities in the text. After manually examining a subset of notes, we roughly estimate that $15\%$ to $25\%$ of entities are not recognized or linked, which may have negatively influenced the prediction model. Some example snippets of clinical notes include:

\begin{small}

``this is a 69 year old man with a history of end stage cardiomyopathy ( nyha class 4 ) and severe chf with an ef of 15 ( ef of 20 on milrinone drip ) as well as severe mr p/w sob , doe , pnd , weight gain of 6lbs in a week , likely due to chf \textbf{exacerbation} . ''

``he has a history of v-tach which responded to amiodarone . patient also has icd in place . respiratory : sob and increased o2 requirement were likely secondary to chf \textbf{exacerbation} and resultant \textbf{pulmonary edema}''

``you were admitted for increasing \textbf{shortness of breath} and oxygen requirements on increasing doses of lasix''

\end{small}

The texts in bold refer to unrecognized entity mentions. Essentially they should be linked to UMLS entities C4086268 (Exacerbation), C0034063 (pulmonary edema) and C0013404 (Dyspnea), respectively. These uncovered entities might indicate the severity of patients' conditions and thus are critical for predicting the readmission risk.

\section{Conclusion}
In this study, we propose a novel graph-based text classification model, i.e., MedText, to predict ICU patient readmission risk, using clinical notes from patients' EHRs. The experiments demonstrate the effectiveness of the method and an updated state-of-the-art performance is observed on the benchmark.

%The model can be easily adapted to other patient outcome prediction tasks, such as mortality prediction.

\begin{acks}
This research has been supported by the Army Research Office (ARO) grant W911NF-21-1-0112 and the NSF grant CNS-1747798 to the IU-CRC Center for Big Learning. This research is also based upon work supported by the Office of the Director of National Intelligence (ODNI), Intelligence Advanced Research Projects Activity (IARPA), via IARPA Contract No. 2019-19051600006 under the Better Extraction from Text Towards Enhanced Retrieval (BETTER) Program. We also would like to thank the IBM-Almaden research group for their support in this work. 
%The views and conclusions contained herein are those of the authors and should not be interpreted as necessarily representing the official policies, either expressed or implied, of ARO, NSF, ODNI, IARPA, the Department of Defense, or the U.S. Government. This document does not contain technology or technical data controlled under either the U.S. International Traffic in Arms Regulations or the U.S. Export Administration Regulations.
\end{acks}

\bibliographystyle{ACM-Reference-Format}
\balance
\bibliography{sample-base}

%\appendix

\end{document}